\documentclass[10pt,twocolumn,letterpaper]{article}

\usepackage{cvpr}
\usepackage{times}
\usepackage{epsfig}
\usepackage{graphicx}
\usepackage{amsmath}
\usepackage{amssymb}

\usepackage{amsmath}
\usepackage{graphics}
\usepackage{graphicx}
\usepackage{amssymb}
\usepackage{algorithm}
\usepackage{algorithmic}
\usepackage{mathrsfs}
\usepackage{booktabs}
\usepackage{bm}
\usepackage{bbm}
\usepackage{float}
\usepackage{url}
\usepackage{amsfonts}
\usepackage[square,sort&compress,numbers]{natbib}
\usepackage{multirow}
\usepackage{textcomp}
\usepackage{supertabular}
\usepackage{array}
\usepackage{caption}
\usepackage{gensymb}

\usepackage[breaklinks=true,bookmarks=false]{hyperref}

\cvprfinalcopy 

\def\cvprPaperID{5946} 

\ifcvprfinal\fi
\begin{document}

\urlstyle{rm}

\newcolumntype{L}[1]{>{\raggedright\arraybackslash}p{#1}}
\newcolumntype{C}[1]{>{\centering\arraybackslash}p{#1}}
\newcolumntype{R}[1]{>{\raggedleft\arraybackslash}p{#1}}

\title{NeuralMagicEye: Learning to See and Understand the Scene Behind an Autostereogram}

\author{
Zhengxia Zou$^1$, \ \ Tianyang Shi$^2$,  \ \ Yi Yuan$^2$,  \ \ Zhenwei Shi$^3$\\
$^1$University of Michigan, Ann Arbor, \ \ $^2$NetEase Fuxi AI Lab, \ \ $^3$Beihang University \\
}

\twocolumn[{%
\renewcommand\twocolumn[1][]{#1}%
\maketitle
\centering{\includegraphics[width=1.0\linewidth]{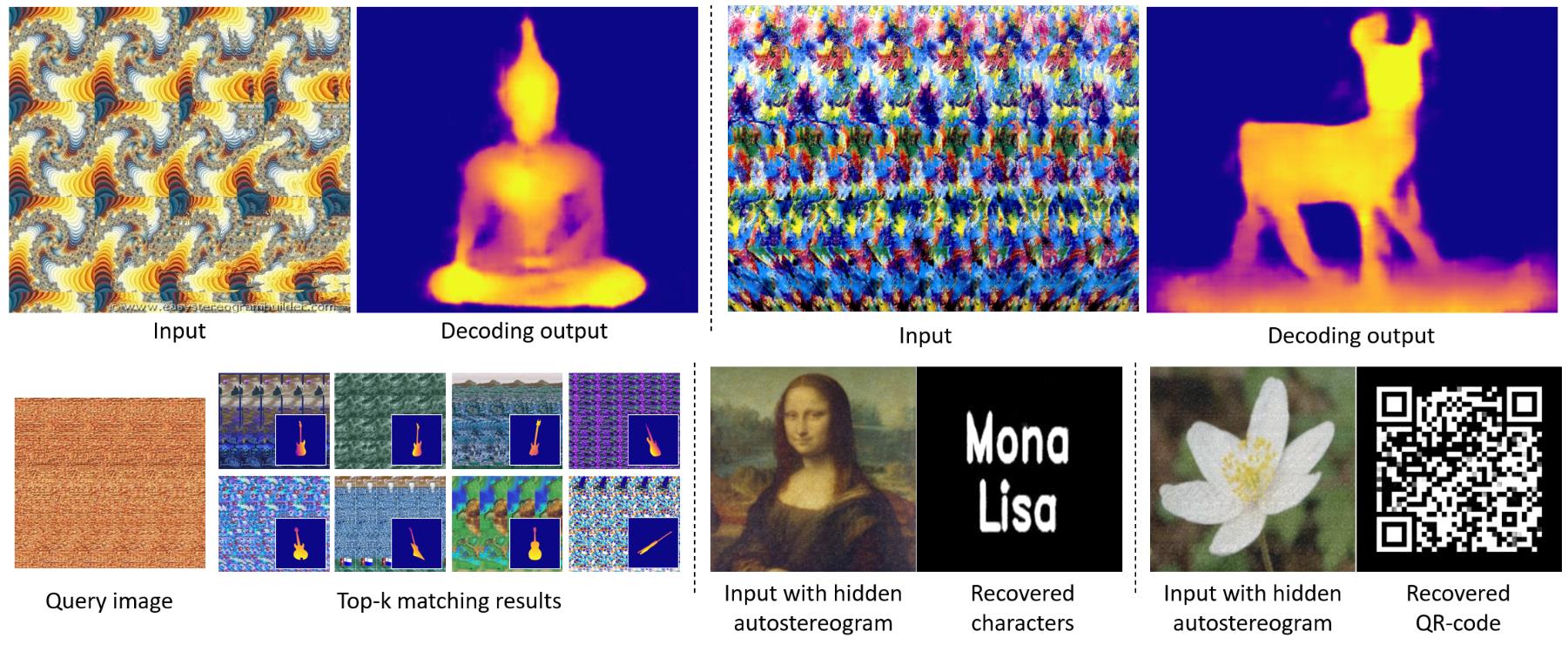}}
\captionof{figure}{\small We propose NeuralMagicEye, a neural network based method for solving autostereograms and understanding the scenes behind. Scan the recovered QR-code with your smartphone and check out for more details about how to view an autostereogram.}
\label{fig:teaser}
\vspace{2em}
}]

\begin{abstract}
\vspace{-0.5em}

An autostereogram, a.k.a. magic eye image, is a single-image stereogram that can create visual illusions of 3D scenes from 2D textures. This paper studies an interesting question that whether a deep CNN can be trained to recover the depth behind an autostereogram and understand its content. The key to the autostereogram magic lies in the stereopsis - to solve such a problem, a model has to learn to discover and estimate disparity from the quasi-periodic textures. We show that deep CNNs embedded with disparity convolution, a novel convolutional layer proposed in this paper that simulates stereopsis and encodes disparity, can nicely solve such a problem after being sufficiently trained on a large 3D object dataset in a self-supervised fashion. We refer to our method as  ``NeuralMagicEye''. Experiments show that our method can accurately recover the depth behind autostereograms with rich details and gradient smoothness. Experiments also show the completely different working mechanisms for autostereogram perception between neural networks and human eyes. We hope this research can help people with visual impairments and those who have trouble viewing autostereograms. Our code is available at \url{https://jiupinjia.github.io/neuralmagiceye/}.
\end{abstract}

\vspace{-1em}
\section{Introduction}

An autostereogram is a picture within a picture where a hidden scene will emerge in 3D when the image viewed correctly~\cite{julesz1986stereoscopic,tyler1990autostereogram,wikiautostereogram}. Autostereograms are also called ``magic eye images'' - viewing autostereograms is like a brain teaser, some people may find it easy, but others may find it very hard even after considerable practice. The key to the magic of autostereograms lies in the stereopsis. Human brain processes depth information from complex mechanisms by matching each set of points in one eye's view with the same set of points in the other eye's view. Autostereograms became part of visual art and widely appear in posters and books after their great popularity in the 1990s~\cite{tyler1990autostereogram,ne1993magic}. It also has attracted much attention for the research of psychophysics and the understanding of the human vision in the past 30 years.

In this paper, we investigate a very interesting question that whether we can train a neural network to recover the hidden depth behind the autostereograms and understand their content. In computer vision and computer graphics, both stereo-vision~\cite{marr1979computational} and autostereogram synthesis~\cite{tyler1990autostereogram,thimbleby1994displaying} are popular and widely studied topics. However, so far, the ``inverse autostereogram'' problem that we will talk about in this paper has rarely been studied. Recovering depth from an autostereogram, which requires detecting disparity from quasi-periodic textures, is totally different from those conventional pixel-wise image prediction tasks, such as image segmentation~\cite{long2015fully,chen2017deeplab} and image super-resolution~\cite{dong2015image}. In conventional pixel-wise prediction tasks, the spatial correspondence in the convolution operation provides an important inductive prior. However, when it comes to predicting the disparity in autostereograms, the loss of spatial correspondences makes it difficult for us to directly apply standard CNNs for such a task. 

The key to our method is called ``disparity convolution'', a novel convolutional layer proposed in this paper, which simulates stereopsis by encoding the feature discrepancy between horizontally neighboring pixels. The disparity convolution can help re-introduce spatial correspondence to the autostereogram decoding process. One of our important conclusions is that deep CNNs with embedded disparity convolution are capable of solving such a problem and greatly improves the decoding accuracy. Another good property of our method is that our training is performed in a pure self-supervised learning fashion, without requiring any manual annotations. To build the self-supervised training loop, we introduce a graphic renderer and an autostereogram generator so that we can continuously generate a large number of autostereograms and the corresponding depth images on the fly. We train our model on ShapeNet~\cite{shapenet2015}, a large 3D object shape dataset, by minimizing the difference between the decoding output and the rendered depth map. Experiments show that after sufficient training under the proposed self-supervised framework, our method can accurately recover the hidden depth behind an autostereogram with rich details and gradient smoothness. When testing on the online autostereograms generated by different graphic engines and those with various degradations (e.g. image blurring, jpg compression, and overexposure), our method also shows very good generalization performance.

Our method also has some potential applications. One is that our method can be naturally applied to the retrieval of autostereograms, where traditional retrieval methods may fail in such cases since the image content (depth) is typically hidden in the textures and can not be perceived directly by those models. Another potential application is digital watermarking. We show that our method is able to encode a hidden layer of depth into an image carrier, and then restore this hidden layer under some noise tolerance. Finally, we hope this research can help people with visual impairments, especially those with amblyopia who have trouble viewing autostereograms.

\begin{figure*}
    \centering{\includegraphics[width=1.0\linewidth]{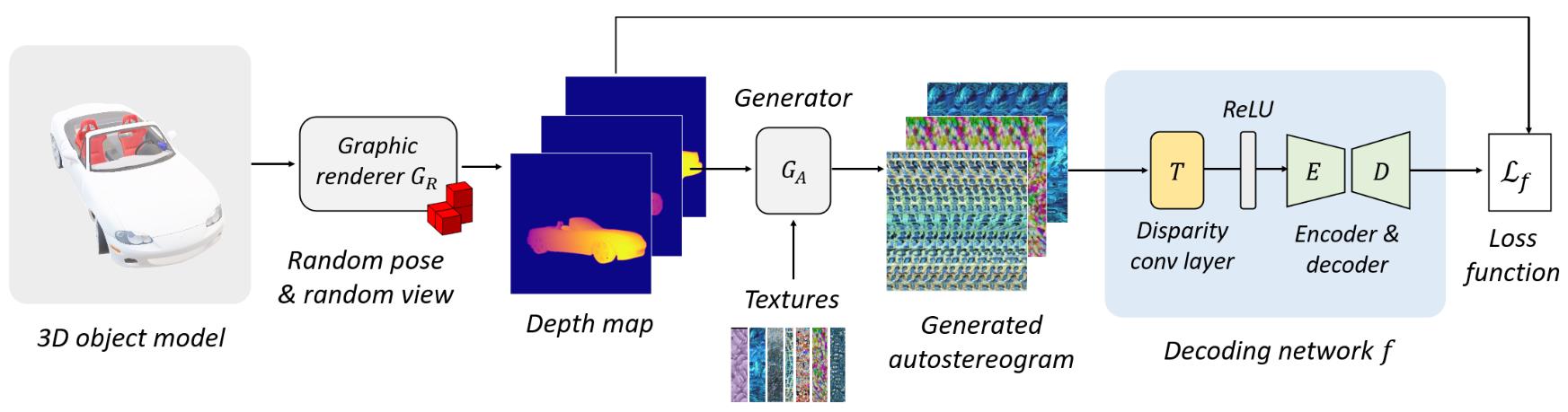}}\\
    \caption{An overview of our method. Our method consists of a graphic renderer $G_R$, an autostereogram generator $G_A$, and a decoding network $f$. We train our method in a self-supervised way by minimizing the difference between its decoding outputs and the rendered depth.}
    \label{fig:overview}
\end{figure*}

\section{Related Work}

The research on autostereograms has a long history and can be traced back to the 1960s~\cite{julesz1962automatic}. Julesz \etal were the first to invent random-dot stereograms and discovered that the impression of depth could arise purely from stereopsis, without relying on other cues such as perspective or contours~\cite{julesz1986stereoscopic}. At the time, scientists believed that depth perception occurred in the eye itself, but now it is proved to be a complex neurological process~\cite{wikiautostereogram}. In the 1990s, Tyler \etal described a way of combining the left and right eye random dots into one picture and invented ``single eye random-dot stereograms'' (later known as autostereograms)~\cite{tyler1990autostereogram}, which do not require the use of special equipment for viewing 3D effect. This new type of stereoscopic image then has attracted the attention of many researchers~\cite{terrell1994behind,thimbleby1994displaying,ninio2007science,nguyen2016stereotag} and quickly became a popular form of visual art~\cite{ne1993magic}.

Despite the extensive research on the synthesis algorithm and applications of autostereograms, the recovery of the hidden depth information in an autostereogram is still a problem that has been rarely studied. Some early works show that self-organizing neural networks can discover disparity in random-dot stereograms~\cite{becker1992self,lee1996nonlinear}, which suggests that neural networks may have great potential for solving this problem. Later, some correlation-based methods were proposed for the decoding of autostereogram in both spatial domain~\cite{kimmel20023d,proudfoot2003autostereogram,kscottz2013} and frequency domain~\cite{hearn2013}. However, the recovered depth image of these methods is often inaccurate and noisy. 
Some of these methods~\cite{proudfoot2003autostereogram} require a manually defined depth searching range, and some 
others~\cite{hearn2013,loch2014} can only extract the outline of the objects and cannot get their exact depth value. Different from all the above approaches, in this paper, we proposed to use deep neural networks to solve this problem and our results show clear advantages over the previous ones.

\begin{figure}
    \centering{\includegraphics[width=1.0\linewidth]{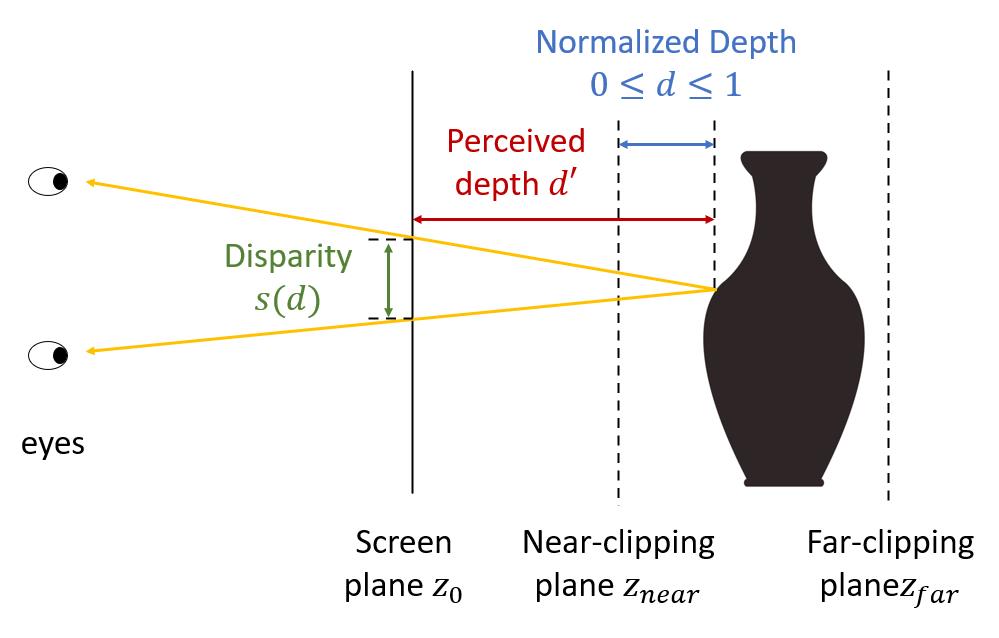}}\\
    \caption{The geometry behind an autostereogram and the principle of autostereogram generation.}
    \label{fig:geometery}
\end{figure}

\section{Methodology}

Fig.~\ref{fig:overview} shows an overview of our method. A complete training pipeline of our method involves three basic modules: 1) a graphic renderer $G_R$ that renders depth images from 3D object models; 2) an autostereogram generator $G_A$ that encodes depth and synthesizes autostereograms; 3) a decoding network $f$ that recovers the depth. In the following, we will first talk about some geometry basis of an autostereogram and then introduce how each module works in detail. 

\subsection{Autostereogram geometry}

The human eye can produce a 3D illusion on 2D patterns that repeats at a certain frequency. In an autostereogram, the depth information is encoded in a series of 2D quasi-periodic textures - when the texture is repeated at a higher frequency, that texture area will appear closer to the screen and vices versa~\cite{tyler1990autostereogram,wikiautostereogram}.  Fig.~\ref{fig:geometery} shows the geometry basis of autostereograms.

Suppose we have a screen placed at the location $z_0$. We set the near-clipping plane and the far-clipping plane of the virtual object to the location $z_{near}$ and $z_{far}$. Within the range of $z_{near}$ to $z_{far}$, each point in the object's surface can be represented as a normalized depth value $d$ ($0\leq d \leq 1$). Through simple geometric calculation, we can establish an approximate relationship between the depth $d$ and disparity $s$ as follows:
\begin{equation}\label{eq:geometry}
    s / s_{d=1} \approx (1-\beta(1-d)),
\end{equation}
where $\beta=(z_{far}-z_{near})/(z_{far}-z_0)$ is a constant. $s_{d=1}$ is the disparity of the far-clipping plane on the screen, as well as the width of the background texture stripes.

Given a depth image and a background stripe, to generate the autostereogram, we first tile the background stripes to fill the entire output image, then scan each pixel in the output image and shift it along the horizontal axis based on the required disparity in (\ref{eq:geometry}). To recover the depth image from an autostereogram, the most straightforward way is to set a small horizontal window for each pixel in the autostereogram and search for correspondence of row segments within a certain searching range. The shifting distance of the best matching window can be thus used to compute the normalized depth $d$ in the depth image. However, as shown in previous research~\cite{proudfoot2003autostereogram}, this method requires manually determining the searching range and may produce noisy and even wrong depth images.

\subsection{Disparity convolution}

In recent years, deep CNNs~\cite{krizhevsky2017imagenet, simonyan2014very, he2016deep} have been extensively used in pixel-wise image prediction tasks, in which the spatial correspondence of the convolution operation provides important inductive priors for such tasks. However, in our method, the networks are trained to learn a mapping from pixels to the disparity. In this case, the spatial correspondence will be lost in most image regions. To address this problem, we propose ``disparity convolution'' by extending a standard convolution with a demodulation operation. The basic idea is to compute the discrepancy of each feature vector in the feature map with its horizontal neighbors and save the values to corresponding feature channels. 

\begin{figure}
    \centering{\includegraphics[width=\linewidth]{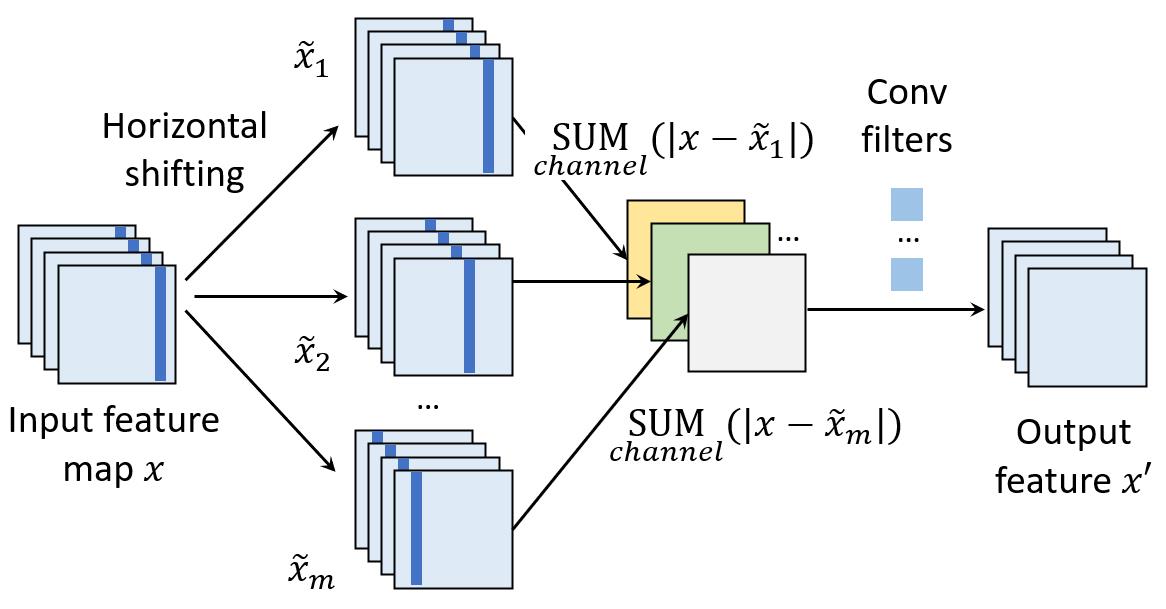}} \\
    \caption{An illustration of the proposed disparity convolution layer.}
    \label{fig:disparity_conv}
\end{figure}

Fig.~\ref{fig:disparity_conv} shows an illustration of the proposed disparity convolution layer. Suppose we have a feature map $x$ with the size of $h \times w \times c$. For each pixel location $(i,j)$ and each channel $k$ in the feature map $x$, we compute the discrepancy between $x(i,j, k)$ and its horizontal neighbor pixel $x(i,j-s,k)$ with $s$-pixel distance. Then we compute a set of difference-maps by sum-up the discrepancy for each spatial location along their channel dimension:
\begin{equation}
    u(i,j,s) = \sum_{k=1}^{c} |x(i,j,k) - x(i,j-s,k)|,
\end{equation}
where $u$ are the generated difference-maps with the size of $h\times w \times m$. $m$ is a pre-defined the distance-bound of horizontal neighbors ($0 < m \leq w$). We finally apply a standard convolution layer on $u$ and produces the output feature map $x^\prime$ as the output of our disparity convolution layer.

To speed up computation, an equivalent but more efficient way of the above calculation is to first circularly shift the feature map along its horizontal axis and then make element-wise subtraction with its input, as shown in Fig.~\ref{fig:disparity_conv}. In PyTorch~\cite{paszke2019pytorch}, this can be further speed-up by using the ``Unfold'' operation and broadcasting, without using any for-loops in the computation. Besides, another good property of the disparity convolution is that it will not introduce any additional parameters compared to a standard convolution layer.

\subsection{Self-supervised training}

We train our autostereogram decoding network under a self-supervised learning paradigm. Given a set of 3D object models (e.g. 3D meshes), we introduce a graphic renderer $G_R$ and for each object model $v$, we use $G_R$ to render a depth image $I_d=G_R(v|\gamma)$ by choosing from a set of random pose and view parameters $\gamma\in \Gamma$, where $\Gamma$ defines the feasible range of the parameters. We then build an autostereogram generator $G_A$, which takes in the rendered depth and a random texture file, and generates an autostereogram image $I_s=G_A(I_d)$. 

For each $I_s$, we feed it into the decoding network $f$ and produce a decoding output $f(I_s)$. Since both the depth image and the generated autostereogram are available during the rendering pipeline, we can use this data pair to effectively train the decoding network $f$ and enforce its output to be similar to the ground truth $I_d$. We train the network $f$ with standard $\ell_2$ pixel regression losses and minimize the following objective function: 
\begin{equation}\label{eq:loss}
\begin{split}
    \mathcal{L}_f &= \mathbb{E}_{v\in V, \gamma\in \Gamma}\{ {\|f(I_s) - I_d\|}_2^2\} \\
    &=\mathbb{E}_{v\in V, \gamma\in \Gamma}\{ {\|f(G_A(G_R(v|\gamma))) - G_R(v| \gamma)\|}_2^2\},
\end{split}
\end{equation}
where $V$ is the dataset of 3D object models, on which our decoder $f$ will be trained. 

\begin{figure*}
    \centering{\includegraphics[width=0.95\linewidth]{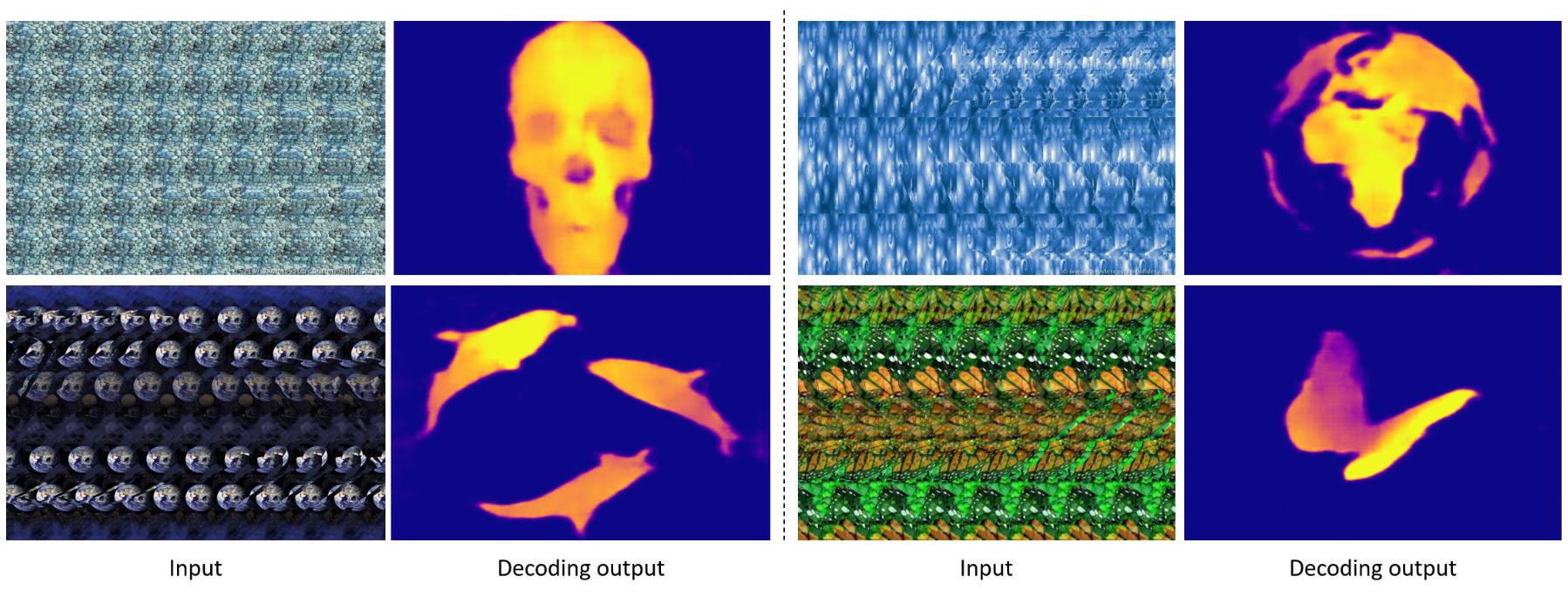}} \\
    \caption{Some online autostereograms and their decoding results by using our method. The autostereograms are generated by different authors using different graphic engines.}
    \label{fig:rst_online}
\end{figure*}

\begin{figure}
    \centering{\includegraphics[width=1.0\linewidth]{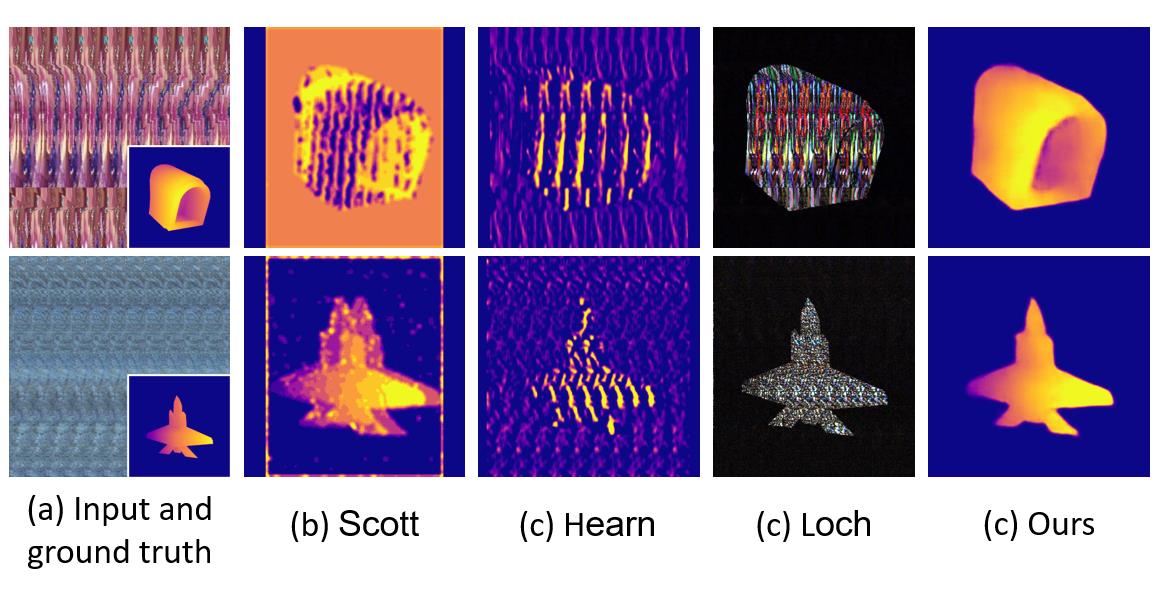}} \\
    \caption{A visual comparison of the autostereogram decoding results by using different methods: ``DeMagicEye'' by K. Scott~\cite{kscottz2013}, ``Magiceye-solver'' by T. Hearn~\cite{hearn2013}, ``Stereogram Viewer'' by F. G. Loch~\cite{loch2014}, and ours.}
    \label{fig:rst_cmp}
\end{figure}

\subsection{Implementation Details}

{\bf Network architecture.} We build our decoding network on top of two popular network architectures: resnet18~\cite{he2016deep} and unet~\cite{ronneberger2015u}. For both architectures, a disparity convolution layer and a ReLU layer are inserted at the input end of the networks. For the resnet18 backbone, we also build a 6-layer upsampling convolution network as its pixel-wise prediction head. For the unet backbone, we tested on different configuration versions with the input image size of 64x64, 128x128, 256x256 pixels respectively. We remove the activation layer from the output layer since in this way we observed a better convergence. In our disparity convolution layer, we set the max shifting distance to 1/4 of the height of the input image. We also tested other configurations such as layer normalization~\cite{ioffe2015batch,ulyanov2016instance} and feature fusion~\cite{lin2017feature}, where more details will be given in our controlled experiment.

{\bf Dataset and image synthesis.} We train our decoding network on ``ShapeNetCore''~\cite{shapenet2015}, a large 3D shape dataset, which covers 55 common object categories with over 50,000 unique 3D models. We use Pyrender~\cite{pyrender} to render 2D depth images of these 3D objects. We set the size of the viewport to 1024x1024 pixels and set yfov to 45\degree. We randomly split the dataset into a training set (90\%) and a testing set (10\%). We also collected 718 texture tiles online (585 for training and 133 for testing) for autostereogram generation. The generated autostereograms are resized to 64x64, 128x128, or 256x256 pixels for training depending on the choice of the network configuration.

{\bf Training details.} We train our decoding network by using Adam optimizer~\cite{kingma2014adam}. We set the batch size to 32, learning rate to 2e-4, and betas to (0.9, 0.999). We set the max epoch number to 100 and reduce the learning rate to its 1/10 after 50 epochs. Data augmentation including random cropping, flipping, rotation, blurring, and jpg compression are performed during the training to enhance the robustness of the decoding network.


\begin{table}[t]\small
\caption{Quantitative comparison of the decoding accuracy of different methods on the autostereograms generated from two datasets: ShapeNet~\cite{shapenet2015} and MNIST~\cite{lecun1998gradient}.}
\centering
\begin{tabular}{r|cc|cc}
\toprule
 & \multicolumn{2}{c}{ShapeNet} & \multicolumn{2}{c}{MNIST} \\
Method & PSNR & SSIM  & PSNR & SSIM \\
\midrule
DeMagicEye~\cite{kscottz2013} & 7.106 & 0.258 & 7.188 & 0.145 \\
Magiceye-solver~\cite{hearn2013} & 10.95 & 0.389 & 11.269 & 0.327 \\
\midrule
Stereo-ResNet (Ours) & 23.12 & \textbf{0.903} & 27.059 & \textbf{0.937} \\
Stereo-UNet (Ours) & \textbf{23.88} & 0.900 & \textbf{27.09} & 0.895 \\
\bottomrule
\end{tabular}
\label{tab:compare_sota}
\end{table}

\begin{figure*}
    \centering{\includegraphics[width=0.98\linewidth]{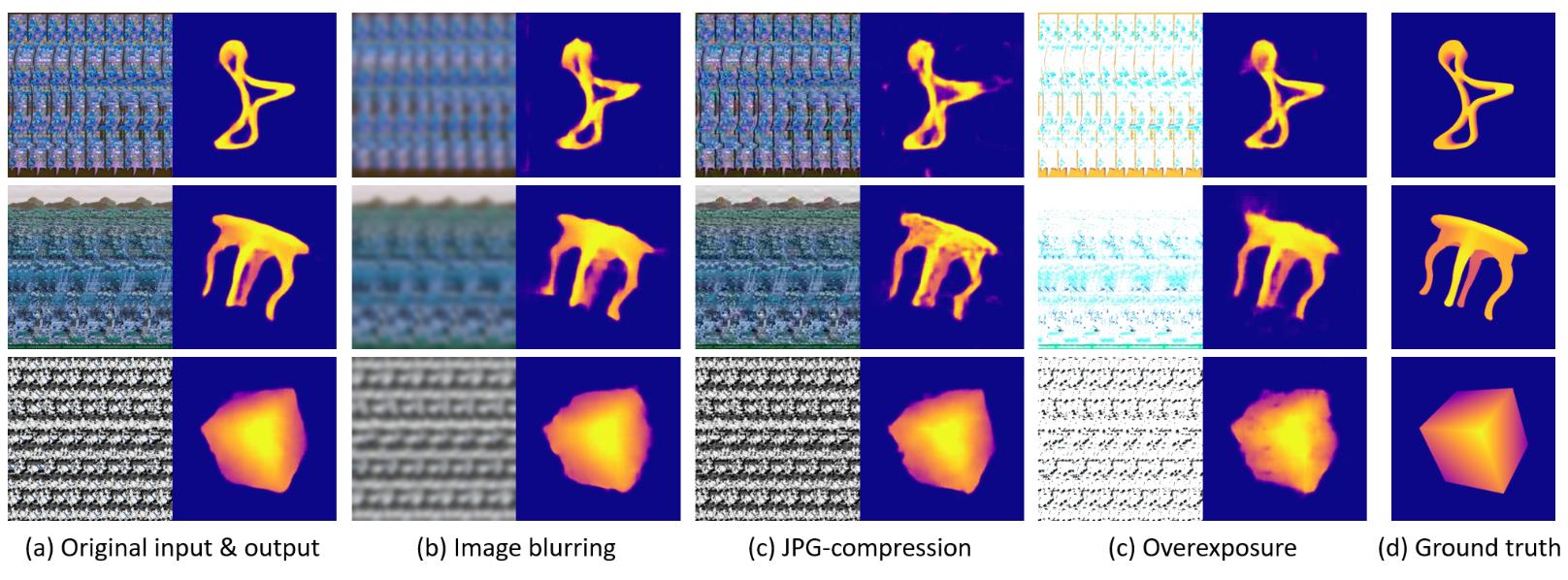}}\\
    \caption{Robustness test of our method on different types of image degradation: image blurring, jpg compression (OpenCV JPEG\_QUALITY=20), and overexposure.}
    \label{fig:robustness}
\end{figure*}

\begin{figure}
    \centering{\includegraphics[width=0.95\linewidth]{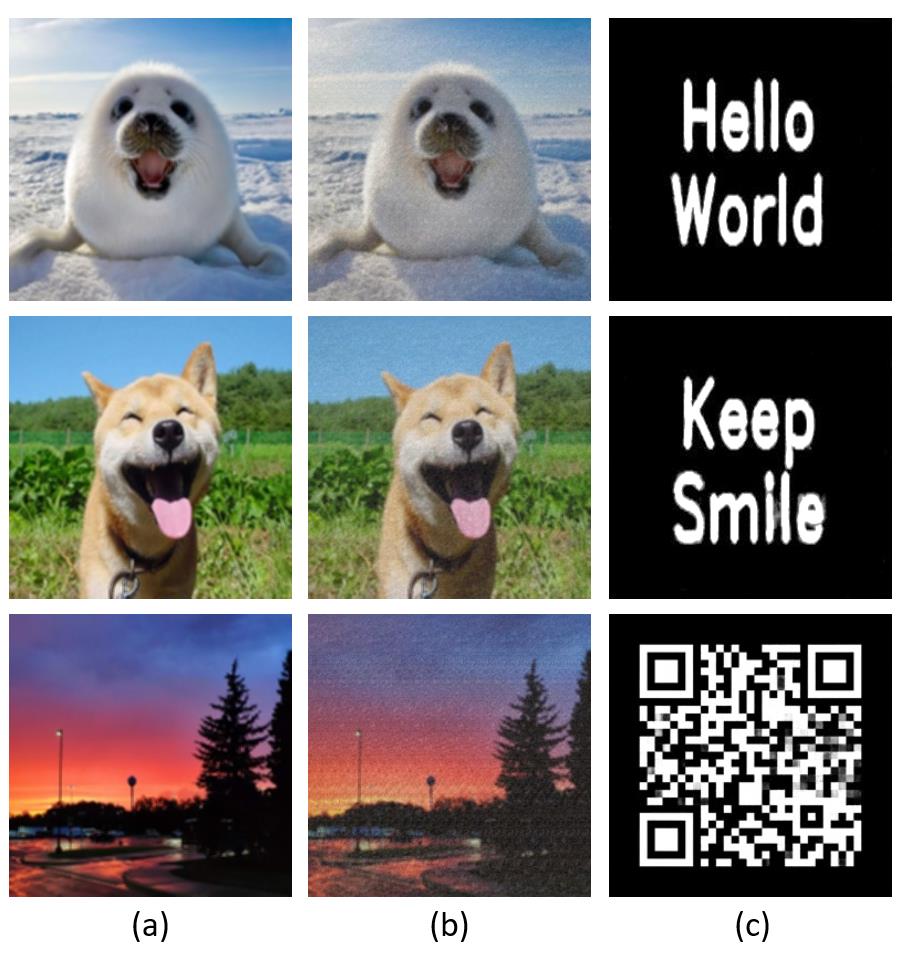}} \\
    \caption{(a) Image carrier. (b) Superimposed images by combining the clean images in (a) and autostereograms in which some characters are encoded. (c) Decoding output from (b). Scan the QR-code with your smartphone and check out more details.}
    \label{fig:wm}
\end{figure}

\section{Experimental Analysis}

\subsection{Decoding autostereograms}

Fig.~\ref{fig:rst_online} shows some online autostereogram images generated by different autostereogram graphic engines and their decoding results by using our method. We compare our method with two open-source autostereogram decoding methods~\cite{kscottz2013,hearn2013}, and a free autostereogram decoding software~\cite{loch2014}. Fig.~\ref{fig:rst_cmp} shows the comparison results. 

In the comparison, we also test on the autostereograms generated from the MNIST dataset~\cite{lecun1998gradient} where the digits are used as depth images. We train our network on MNIST for 10 epochs. Table~\ref{tab:compare_sota} shows their decoding accuracy. We use PSNR and SSIM~\cite{wang2004image} as metrics of recovery accuracy. Both the decoding result and the ground truth depth are normalized to [0, 1] before the accuracy is computed. We can see our method produces much better results in terms of both visual quality and quantitative scores.

\subsection{Robustness}

To test the robustness of our decoding network, we apply different degradations to the autostereograms, including image blurring, jpg compression, and overexposure. Fig.~\ref{fig:robustness} shows the degraded input and the recovered depth. We can see that although many high-frequency details are lost after degradation, the depth can be still nicely recovered. This shows that the neural network has successfully learned to perform disparity prediction and information loss recovery at the same time. Besides, the results of overexposure show that our method is not sensitive to global intensity changes.

\subsection{Digital watermarking}

In this experiment, we investigate whether neural networks are capable of recovering depth from a carrier image in which the autostereogram is embedded as hidden watermarks. We first generate a set of autostereograms based on random characters and QR-codes. The pixel values in the character images and QR-codes are recorded as the depth value in the autostereograms. We then train our decoder on a set of superimposed images $I_w$. These images are generated as a linear combination of a background image carrier $I_b$ and an autostereogram $I_s$: 
\begin{equation}
    I_w = \alpha I_s + (1 - \alpha) I_b.
\end{equation}
We set $\alpha$ as a random number within [0.1, 0.9] during training and set $\alpha=0.2$ during testing. It is worth noting that this task is very challenging since the depth is first encoded as disparities within the textures and then mixed with the image carrier. The decoder is thus trained to deal with a ``double recovery'' problem of the hidden depth information. Fig.~\ref{fig:wm} shows a group of clean images (image carrier), superimposed images, and their decoding outputs. We can see that although the autostereograms are barely visible in the superimposed images, the hidden information can be still clearly recovered by our decoder.

\begin{figure}
    \centering{\includegraphics[width=\linewidth]{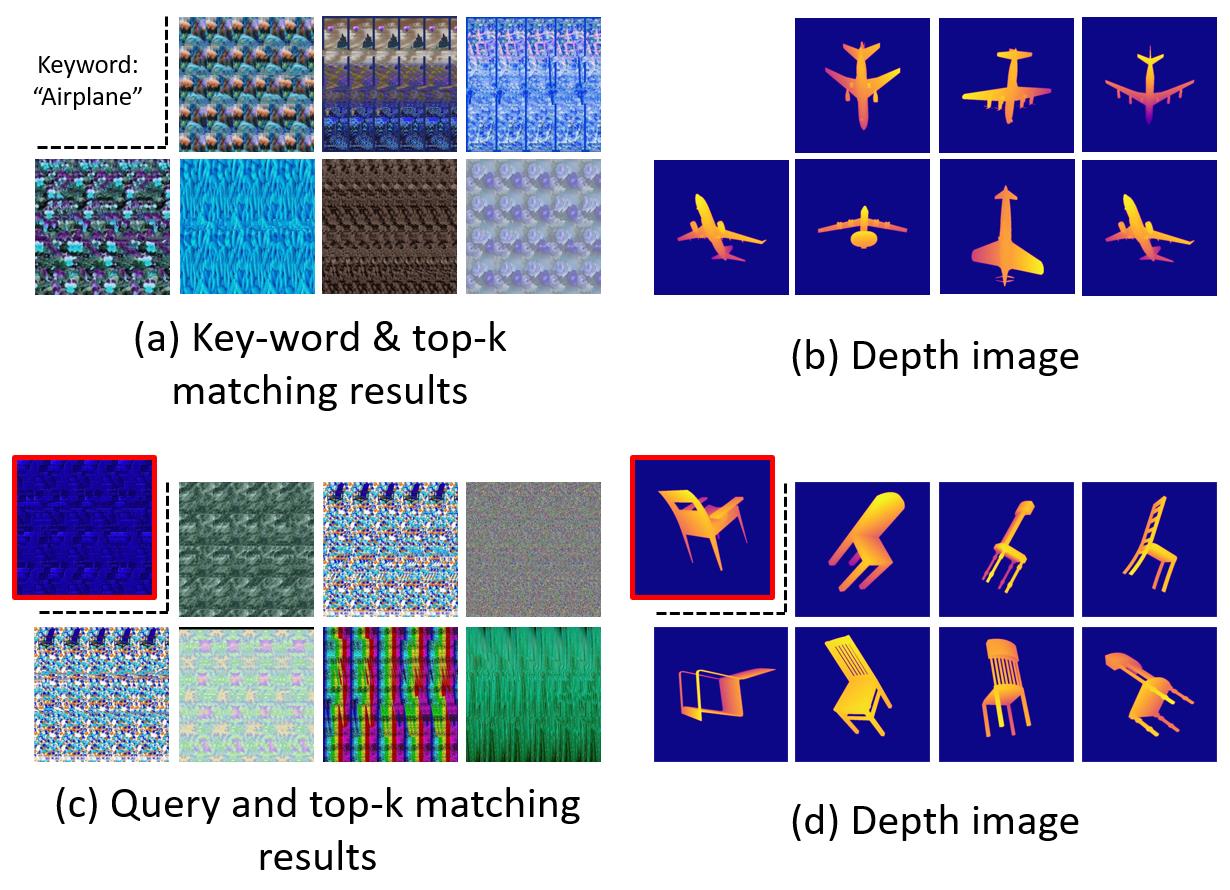}} \\
    \caption{Some image retrieval results by using our method. We test on two retrieval modes: (a)-(b) keywords based retrieval, and (c)-(d) query image based retrieval.}
    \label{fig:retrieval}
\end{figure}

\subsection{Autostereogram retrieval}

Our method can be also applied to autostereogram retrieval. To complete such a task, the model has to learn to understand the semantics behind the autostereograms. We, therefore, build an autostereogram recognition network by replacing the upsampling convolution head in our decoding network with a fully connected layer to predict the class probability of the input. We test on two retrieval modes: retrieval by keywords and by query image. In the keyword retrieval mode, we iterate through all the autostereograms in the database (our testing set) and select the top-k matching results based on the prediction class probability. In the query image retrieval mode, the top-k matching results are selected by computing the feature distance between the query and the matching images. Fig.~\ref{fig:retrieval} shows the retrieval results of the above two modes. We can see although the top-k matching results have very different texture appearance, they share the same semantics in their depth images.

\begin{figure}
    \centering{\includegraphics[width=\linewidth]{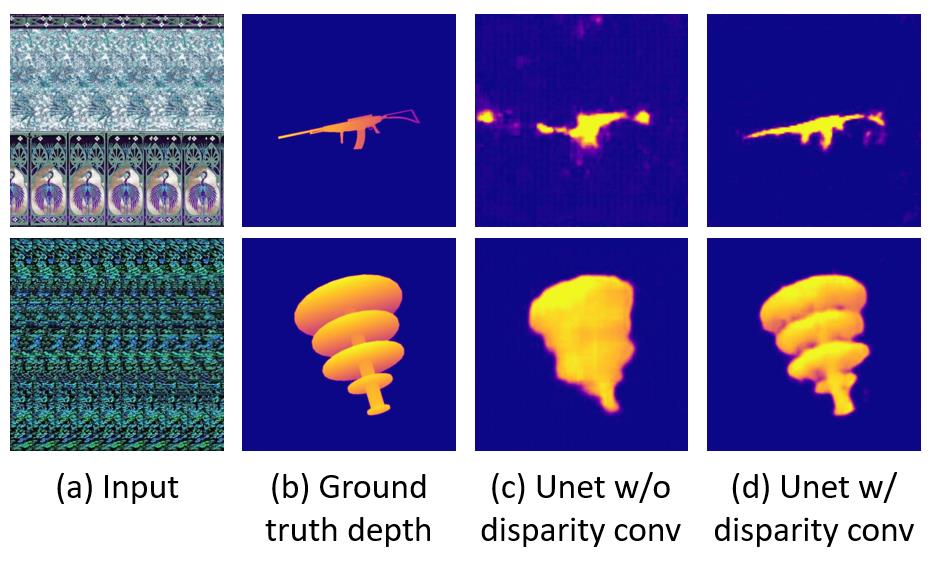}} \\
    \caption{A visual comparison of the decoding output of our method w/ and w/o using disparity convolution.}
    \label{fig:cmp_disparity_conv}
\end{figure}

\begin{table*}\small
\caption{Controlled experiments on the influence of each component in our decoding model, including the choice of different backbones, layer normalization, disparity convolution, and feature fusion. The scores are computed as the mean pixel accuracy of the decoding outputs on the autostereograms generated from two datasets (ShapeNet~\cite{shapenet2015} and MNIST~\cite{lecun1998gradient}).}
\centering
\begin{tabular}{l|cccc|cc|cc}
\toprule
\multicolumn{1}{c}{} &
\multicolumn{4}{c}{Configuration} &
\multicolumn{2}{c}{ShapeNet Decode} &
\multicolumn{2}{c}{MNIST Decode} \\
\cmidrule{2-9}
& Backbone & Normalization & DisparityConv & FeatFusion & PSNR & SSIM  & PSNR & SSIM  \\
\midrule
Res18F & resnet18 & none &   &   & 18.16 & 0.777 & 17.83 & 0.787 \\

Res18F (BN) & resnet18 & batch &   &   & 19.92 & 0.841 & 18.99 & 0.808 \\

Stereo-Res18F (BN) & resnet18 & batch & $\checkmark$  &   & 22.63 & 0.900 & 25.31 & 0.919 \\

Res18F (IN) & resnet18 & instance &   &   & 20.34 & 0.851 & 18.84 & 0.812 \\

Stereo-Res18F (IN) & resnet18 & instance & $\checkmark$  &   & 22.53 & 0.896 & 24.34 & 0.905 \\

Res18F (BN) + FF & resnet18 & batch &   & $\checkmark$  & 20.46 & 0.863 & 19.49 & 0.802 \\

Stereo-Res18F (BN) + FF & resnet18 & batch & $\checkmark$  & $\checkmark$  & 23.12 & 0.903 & 27.06 & 0.937 \\

Res18F (IN) + FF & resnet18 & instance &   & $\checkmark$  & 20.69 & 0.865 & 20.08 & 0.825 \\

Stereo-Res18F (IN) + FF & resnet18 & instance & $\checkmark$  & $\checkmark$  & 23.17 & 0.909 & 26.94 & 0.937 \\

\midrule
UNet & unet128 & none &   &  -- & 20.48 & 0.822 & 19.68 & 0.820 \\

UNet (BN) & unet128 & batch &   & --  & 20.85 & 0.803 & 21.32 & 0.827 \\

Stereo-UNet (BN) & unet128 & batch & $\checkmark$  & --  & 23.88 & 0.900 & 27.09 & 0.895 \\

UNet (IN) & unet128 & instance &   &  -- & 20.76 & 0.764 & 22.08 & 0.830 \\

Stereo-UNet (IN) & unet128 & instance & $\checkmark$  & --  & 23.70 & 0.842 & 27.26 & 0.915 \\

\bottomrule
\end{tabular}%
\label{tab:ablation_decoding}%
\end{table*}%

\begin{table*}\small
\caption{Controlled experiments on the influence of each component in our recognition model, including the choice of different backbones, layer normalization, disparity convolution. The scores are computed as the classification accuracy on the autostereograms generated from two datasets (ShapeNet~\cite{shapenet2015} and MNIST~\cite{lecun1998gradient}). We also evaluate an upper-bound model (*Upperbound Res18) where the model is trained and tested directly on the depth images.}
\centering
\begin{tabular}{l|ccc|c|c}
\toprule
\multicolumn{1}{c}{} &
\multicolumn{3}{c}{Configuration} &
\multicolumn{1}{c}{} &
\multicolumn{1}{c}{} \\
& Backbone & Normalization & DisparityConv  & ShapeNet & MNIST  \\
\midrule
Res18 & resnet18 & none &   & -- & 93.30\% \\

Res18 (BN) & resnet18 & batch &   & 48.06\% & 96.23\%  \\

Stereo-Res18 (BN) & resnet18 & batch & $\checkmark$  & 66.81\% & 98.68\% \\

Res18F (IN) & resnet18 & instance &    & 47.60\% & 96.74\% \\

Stereo-Res18 (IN) & resnet18 & instance & $\checkmark$  & 64.60\% & 98.73\% \\

\midrule
*Upperbound (Res18) & resnet18 & batch & --  & 77.26\% & 99.41\% \\

\bottomrule
\end{tabular}%
\label{tab:ablation_classfication}%
\end{table*}%

\subsection{Controlled experiments}

We designed a series of controlled experiments to comprehensively study the influence of different technical components in our method. These components include the use of our disparity convolution, layer normalization, and feature pyramid fusion. All the controlled experiments are conducted on both ShapeNet and MNIST dataset. We evaluate the recognition accuracy and pixel-level decoding accuracy of our networks with different architecture configurations, as shown in Table~\ref{tab:ablation_decoding} and Table~\ref{tab:ablation_classfication}. Since we do not need a pixel-wise prediction head in our classification task, in Table~\ref{tab:ablation_classfication}, we only evaluate the resnet18 architecture with the disparity convolution and layer normalization. Some useful conclusions on each of these components can be drawn as follows.

\begin{figure}
    \centering{\includegraphics[width=\linewidth]{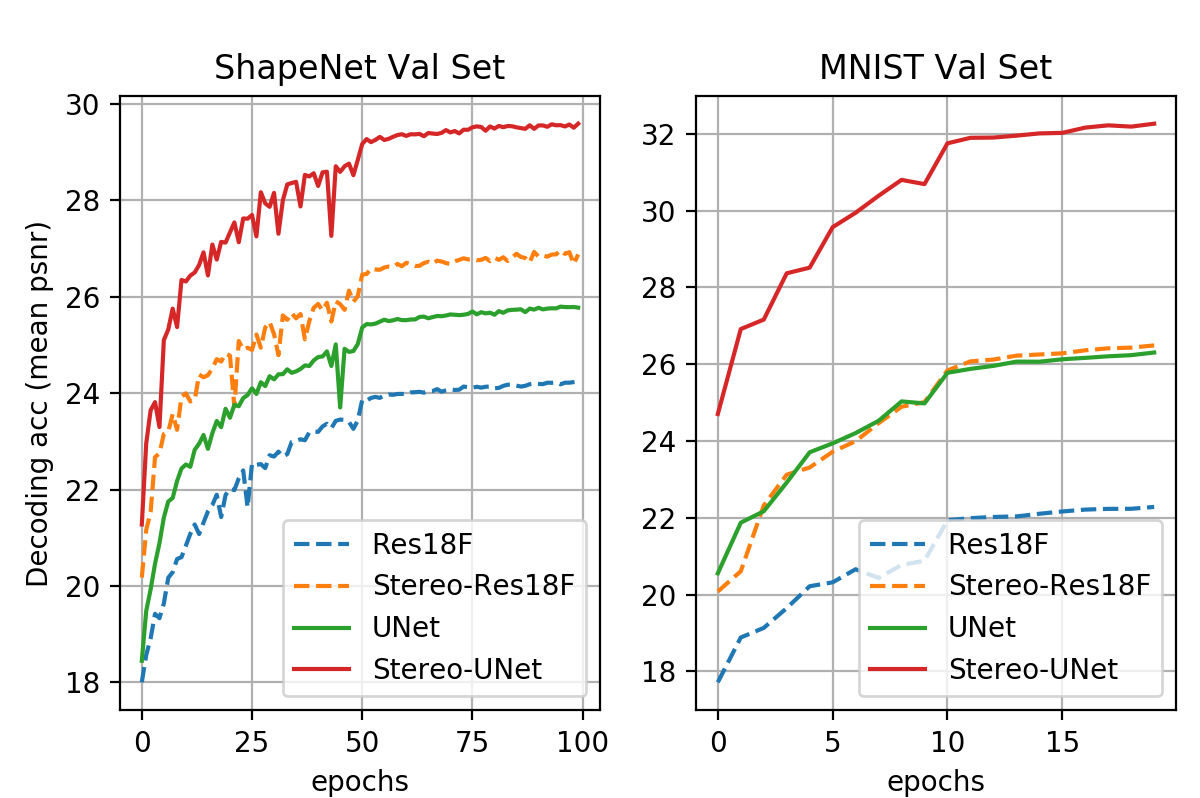}} \\
    \caption{Validation accuracy on different training epochs with (Stereo-Res18F, Stereo-UNet) and without (Res18F, UNet) using disparity convolutions.}
    \label{fig:curve_disparity}
\end{figure}

\begin{figure}
    \centering{\includegraphics[width=\linewidth]{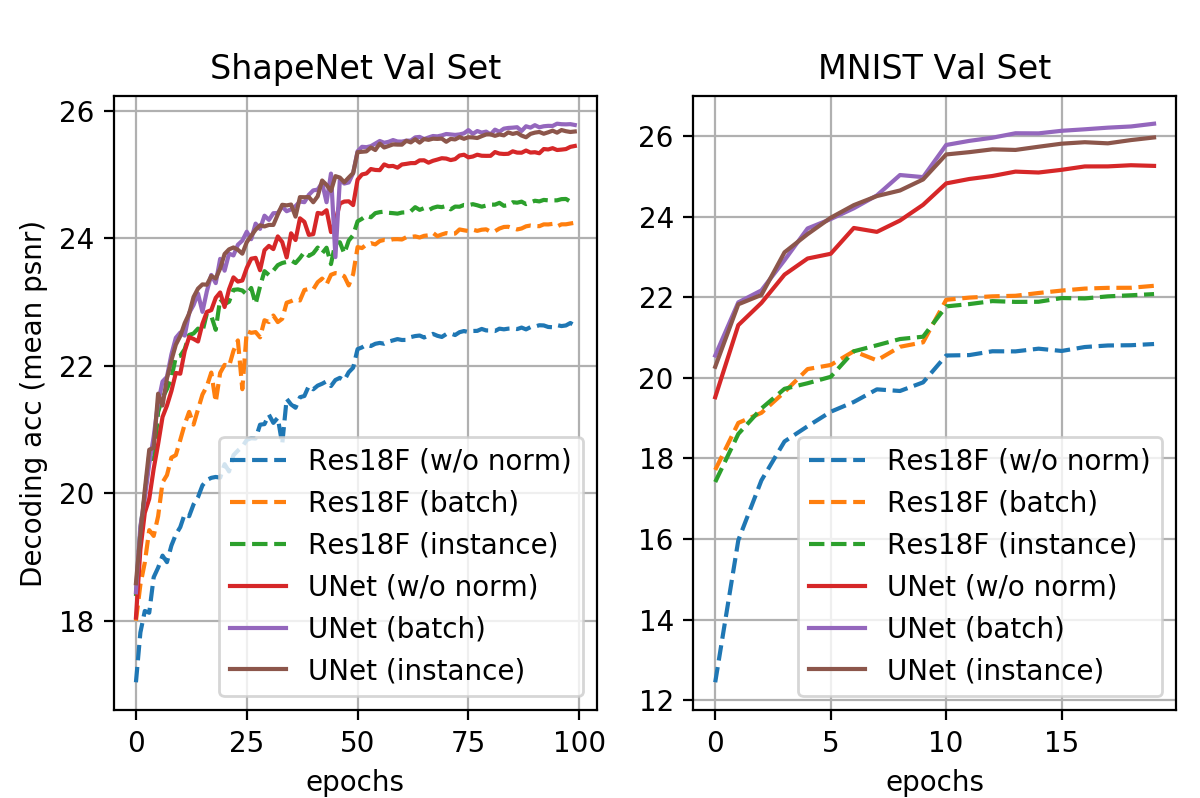}} \\
    \caption{Validation accuracy during the training with different layer normalization methods.}
    \label{fig:curve_norm}
\end{figure}

\begin{figure}
    \centering{\includegraphics[width=\linewidth]{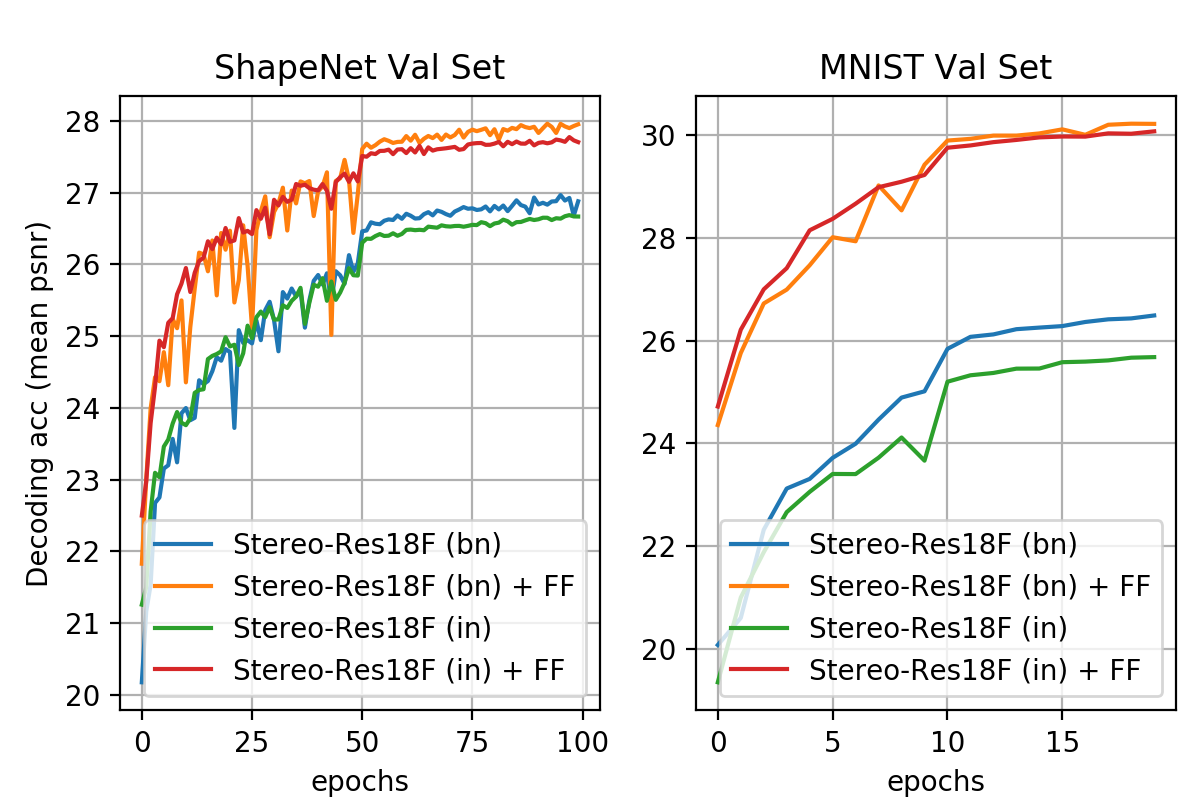}} \\
    \caption{Validation accuracy on different training epochs with (+FF) and without using feature fusion in the pixel-wise prediction head.}
    \label{fig:curve_fpn}
\end{figure}

\textbf{Disparity convolution.} Our experiments show that the disparity convolution plays a crucial role in both decoding and classification tasks. From row 6-9, 11-14 of Table~\ref{tab:ablation_decoding} and row 2-5 of Table~\ref{tab:ablation_classfication}, we can see that when we remove the disparity convolution layers from our best models, we see a significant accuracy drop on both the decoding task (resnet PSNR -2.48 on ShapeNet, -7.57 on MNIST; unet PSNR -3.03 on ShapeNet, -5.18 on MNIST) and the classification task (resnet acc -18.75\% on ShapeNet; -1.99\% on MNIST). Besides, from row 2-5 of Table~\ref{tab:ablation_decoding}, we can see the integrating of the disparity convolution in some other sub-optimal models also brings noticeable improvement on their decoding/classification accuracy (resnet PSNR +2.71 on ShapeNet, +6.32 on MNIST). In Fig.~\ref{fig:curve_disparity}, we plot the validation accuracy on different training epochs.\footnote{It should be noted that in Fig.~\ref{fig:curve_disparity}, \ref{fig:curve_norm} and \ref{fig:curve_fpn}, the validation accuracy is computed on 128x128 images, while in Table~\ref{tab:ablation_decoding}, the final testing accuracy is computed on 512x512 images. That is why there is an accuracy gap between these two groups of results.} We can see that the integration of the disparity convolution not only brings a higher accuracy but also a much faster convergence speed in both resnet and unet architectures.

\textbf{Normalization.} In this experiment, we first start from training a baseline model without using any layer normalization. Then we try two popular normalization methods - batch normalization~\cite{ioffe2015batch} and instance normalization~\cite{ulyanov2016instance} and compare with the baselines. The impact on the resnet and unet architectures are both evaluated. From row 1, 2, 4, 10, 11, 13 of Table~\ref{tab:ablation_decoding}, and row 1, 2, 4 of Table~\ref{tab:ablation_classfication}, we can see that the layer normalization are important for both tasks. For example, batch normalization brings noticeable accuracy improvement on the decoding (resnet PSNR +1.76 on ShapeNet, +1.16 on MNIST; unet PSNR +0.37 on ShapeNet, +1.64 on MNIST) and classification tasks (resnet acc +2.93\% on MNIST). Particularly, we notice that when there is no normalization used in the ShapeNet classification task, the model will not converge during the training. In Fig.~\ref{fig:curve_norm}, we show the validation accuracy w/ and w/o using layer normalization on different training epochs.

\textbf{Feature fusion.} In conventional pixel-wise prediction tasks, feature fusion between layers of different depth and resolutions are crucial for generating high-resolution outputs. The representatives of such an idea include the layer skip connection in the unet~\cite{ronneberger2015u} and feature pyramid networks in the object detection networks~\cite{lin2017feature}.  From row 3, 5, 7, 9 of Table~\ref{tab:ablation_decoding}, we can see that the feature pyramid fusion can bring a minor accuracy improvement (resnet PSNR +0.49 on ShapeNet, +1.75 on MNIST) on our sub-optimal models. When testing on our baseline models (w/o disparity conv), the feature fusion can also bring similar improvement on the decoding accuracy (resnet PSNR +0.54 on ShapeNet, +0.5 on MNIST). These results suggest that in autostereogram decoding, although the spatial correspondences are much weaker than other image pixel prediction tasks, feature fusion can also slightly improve the accuracy of the decoding outputs. In Fig.~\ref{fig:curve_fpn}, we show the validation accuracy (PSNR) w/ and w/o using feature fusion on different training epochs.

\begin{figure}
    \centering{\includegraphics[width=0.95\linewidth]{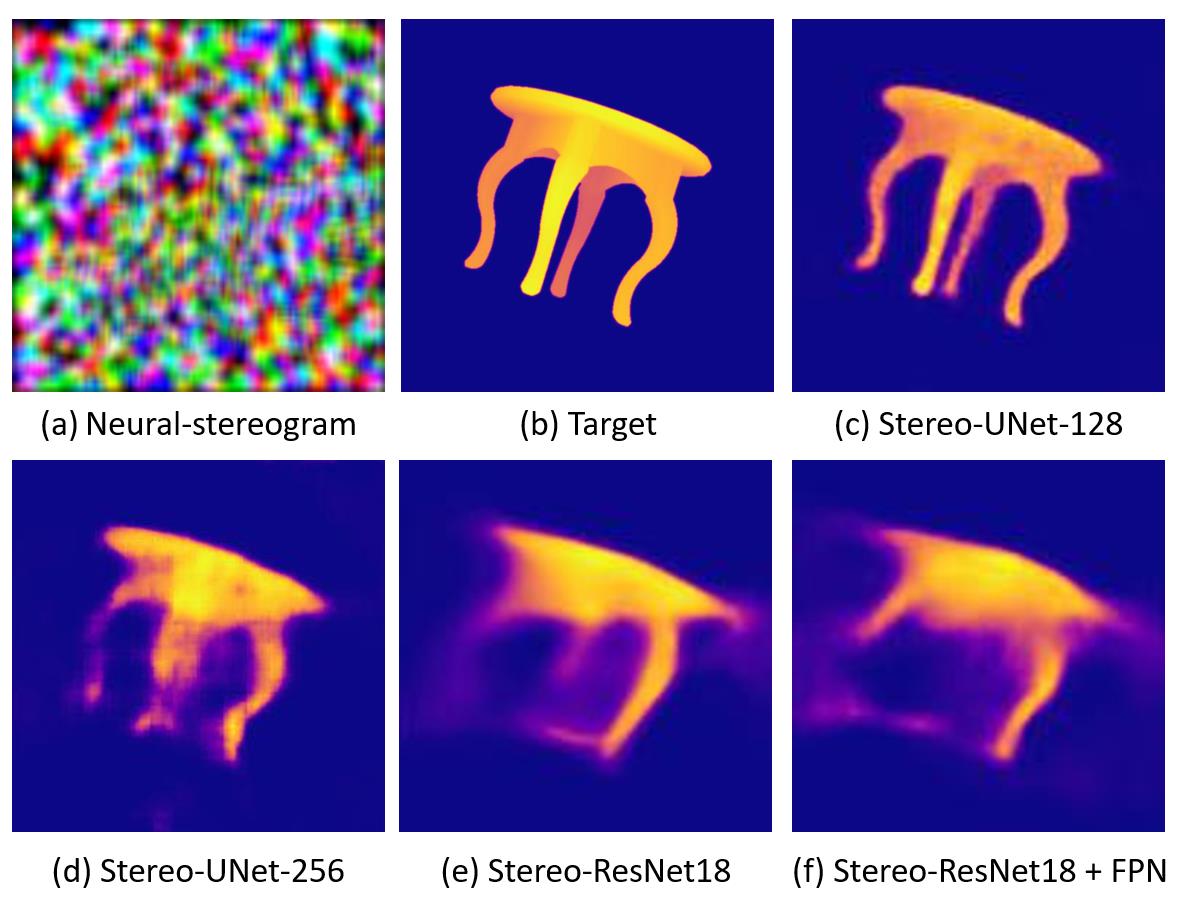}} \\
    \caption{(a) An ``neural autostereogram'' generated by our unet-128 network that minimizes the difference between the target depth (b) and its decoding output (c). (d)-(f) Decoding outputs of the autostereogram in (a) with different network architectures. We found surprisingly that although the depth information in the generated autostereogram (a) cannot be perceived by human eyes, the depth can be nicely recovered by neural networks with different architectures. This experiment suggests that different neural networks may follow a certain commonality when processing autostereograms, but the mechanism behind may be completely different from that of human eye stereopsis.}
    \label{fig:neural_stereogram}
\end{figure}

\subsection{Neural autostereogram}

We finally investigated an interesting question that given a depth image, what an ``optimal autostereogram'' should look like in the eyes of a decoding network. The study of this question may help us understand the working mechanism of neural networks for autostereogram perception. To generate the ``optimal autostereogram'', we run gradient descent on the input end of our decoding network and minimize the difference between its output and the reference depth image. In Fig.~\ref{fig:neural_stereogram} (a), we show the generated ``optimal autostereogram'' on our unet decoding network. We named it ``neural autostereogram''. In Fig.~\ref{fig:neural_stereogram} (b)-(c) we show the reference depth and the decoding output. 

An interesting thing we observed during this experiment is that, although the decoding output of the network is already very similar to the target depth image, however, human eyes still cannot perceive the depth hidden in this neural autostereogram. Also, there are no clear periodic patterns in this image, which is very different from those autostereograms generated by using graphic engines. More surprisingly, when we feed this neural autostereogram to other decoding networks with very different architectures, we found that these networks can miraculously perceive the depth correctly. To confirm that it is not accidental, we also tried different image initialization and smooth constraints but still have similar observations. Fig.~\ref{fig:neural_stereogram} (d)-(f) show the decoding results on this image by using other different decoding networks. This experiment suggests that neural networks and the human eye may use completely different ways for stereogram perception. The mechanism and properties behind the neural autostereograms are still open questions and need further study. 

\section{Conclusion}

We propose ``NeuralMagicEye'', a deep neural network based method for perceiving and understanding the scenes behind autostereograms. We train our networks on a large 3D object dataset under a self-supervised learning fashion and the results show good robustness and much better decoding accuracy than other methods. We also show that the disparity convolution, a new form of convolution proposed in this paper, can greatly improve the autostereogram decoding accuracy of deep CNNs. Controlled experiments suggest the effectiveness of our design. Finally, we show that the way a neural network perceiving depth in an autostereogram is different from that of human eyes. The deeper mechanism behind this needs to be further explored.

{\small
\bibliographystyle{ieee_fullname}
\bibliography{egbib}

\begin{thebibliography}{10}\itemsep=-1pt

\bibitem{becker1992self}
Suzanna Becker and Geoffrey~E Hinton.
\newblock Self-organizing neural network that discovers surfaces in random-dot
  stereograms.
\newblock {\em Nature}, 355(6356):161--163, 1992.

\bibitem{shapenet2015}
Angel~X. Chang, Thomas Funkhouser, Leonidas Guibas, Pat Hanrahan, Qixing Huang,
  Zimo Li, Silvio Savarese, Manolis Savva, Shuran Song, Hao Su, Jianxiong Xiao,
  Li Yi, and Fisher Yu.
\newblock {ShapeNet: An Information-Rich 3D Model Repository}.
\newblock Technical Report arXiv:1512.03012 [cs.GR], Stanford University ---
  Princeton University --- Toyota Technological Institute at Chicago, 2015.

\bibitem{chen2017deeplab}
Liang-Chieh Chen, George Papandreou, Iasonas Kokkinos, Kevin Murphy, and Alan~L
  Yuille.
\newblock Deeplab: Semantic image segmentation with deep convolutional nets,
  atrous convolution, and fully connected crfs.
\newblock {\em IEEE transactions on pattern analysis and machine intelligence},
  40(4):834--848, 2017.

\bibitem{dong2015image}
Chao Dong, Chen~Change Loy, Kaiming He, and Xiaoou Tang.
\newblock Image super-resolution using deep convolutional networks.
\newblock {\em IEEE transactions on pattern analysis and machine intelligence},
  38(2):295--307, 2015.

\bibitem{ne1993magic}
NE~Thing Enterprises.
\newblock {\em The Magic Eye, Volume I: A New Way of Looking at the World}.
\newblock Andrews McMeel Publishing, 1993.

\bibitem{loch2014}
{Francis G. Loch}.
\newblock Stereogram viewer v0.3a, 2014.
\newblock [Online; accessed 12-30-2020].

\bibitem{he2016deep}
Kaiming He, Xiangyu Zhang, Shaoqing Ren, and Jian Sun.
\newblock Deep residual learning for image recognition.
\newblock In {\em Proceedings of the IEEE conference on computer vision and
  pattern recognition}, pages 770--778, 2016.

\bibitem{ioffe2015batch}
Sergey Ioffe and Christian Szegedy.
\newblock Batch normalization: Accelerating deep network training by reducing
  internal covariate shift.
\newblock {\em arXiv preprint arXiv:1502.03167}, 2015.

\bibitem{julesz1986stereoscopic}
Bela Julesz.
\newblock Stereoscopic vision.
\newblock {\em Vision Research}, 26(9):1601--1612, 1986.

\bibitem{julesz1962automatic}
Bela Julesz and Joan~E Miller.
\newblock Automatic stereoscopic presentation of functions of two variables.
\newblock {\em Bell System Technical Journal}, 41(2):663--676, 1962.

\bibitem{kscottz2013}
{Katherine Scott}.
\newblock Solving autostereograms aka magic eyes, 2013.
\newblock [Online; accessed 12-30-2020].

\bibitem{kimmel20023d}
Ron Kimmel.
\newblock 3d shape reconstruction from autostereograms and stereo.
\newblock {\em Journal of Visual Communication and Image Representation},
  13(1-2):324--333, 2002.

\bibitem{kingma2014adam}
Diederik~P Kingma and Jimmy Ba.
\newblock Adam: A method for stochastic optimization.
\newblock {\em arXiv preprint arXiv:1412.6980}, 2014.

\bibitem{krizhevsky2017imagenet}
Alex Krizhevsky, Ilya Sutskever, and Geoffrey~E Hinton.
\newblock Imagenet classification with deep convolutional neural networks.
\newblock {\em Communications of the ACM}, 60(6):84--90, 2017.

\bibitem{lecun1998gradient}
Yann LeCun, L{\'e}on Bottou, Yoshua Bengio, and Patrick Haffner.
\newblock Gradient-based learning applied to document recognition.
\newblock {\em Proceedings of the IEEE}, 86(11):2278--2324, 1998.

\bibitem{lee1996nonlinear}
Christopher~W Lee and Bruno~A Olshausen.
\newblock A nonlinear hebbian network that learns to detect disparity in
  random-dot stereograms.
\newblock {\em Neural computation}, 8(3):545--566, 1996.

\bibitem{lin2017feature}
Tsung-Yi Lin, Piotr Doll{\'a}r, Ross Girshick, Kaiming He, Bharath Hariharan,
  and Serge Belongie.
\newblock Feature pyramid networks for object detection.
\newblock In {\em Proceedings of the IEEE conference on computer vision and
  pattern recognition}, pages 2117--2125, 2017.

\bibitem{long2015fully}
Jonathan Long, Evan Shelhamer, and Trevor Darrell.
\newblock Fully convolutional networks for semantic segmentation.
\newblock In {\em Proceedings of the IEEE conference on computer vision and
  pattern recognition}, pages 3431--3440, 2015.

\bibitem{marr1979computational}
David Marr and Tomaso Poggio.
\newblock A computational theory of human stereo vision.
\newblock {\em Proceedings of the Royal Society of London. Series B. Biological
  Sciences}, 204(1156):301--328, 1979.

\bibitem{pyrender}
{Matthew Matl}.
\newblock Pyrender, 2019.
\newblock [Online; accessed 12-30-2020].

\bibitem{nguyen2016stereotag}
Minh Nguyen and Albert Yeap.
\newblock Stereotag: A novel stereogram-marker-based approach for augmented
  reality.
\newblock In {\em 2016 23rd International Conference on Pattern Recognition
  (ICPR)}, pages 1059--1064. IEEE, 2016.

\bibitem{ninio2007science}
Jacques Ninio.
\newblock The science and craft of autostereograms.
\newblock {\em Spatial vision}, 21(1-2):185--200, 2007.

\bibitem{paszke2019pytorch}
Adam Paszke, Sam Gross, Francisco Massa, Adam Lerer, James Bradbury, Gregory
  Chanan, Trevor Killeen, Zeming Lin, Natalia Gimelshein, Luca Antiga, et~al.
\newblock Pytorch: An imperative style, high-performance deep learning library.
\newblock In {\em Advances in neural information processing systems}, pages
  8026--8037, 2019.

\bibitem{proudfoot2003autostereogram}
Kekoa Proudfoot.
\newblock An autostereogram decoder, 2003.

\bibitem{ronneberger2015u}
Olaf Ronneberger, Philipp Fischer, and Thomas Brox.
\newblock U-net: Convolutional networks for biomedical image segmentation.
\newblock In {\em International Conference on Medical image computing and
  computer-assisted intervention}, pages 234--241. Springer, 2015.

\bibitem{simonyan2014very}
Karen Simonyan and Andrew Zisserman.
\newblock Very deep convolutional networks for large-scale image recognition.
\newblock {\em arXiv preprint arXiv:1409.1556}, 2014.

\bibitem{terrell1994behind}
Maria~S Terrell and Robert~E Terrell.
\newblock Behind the scenes of a random dot stereogram.
\newblock {\em The American Mathematical Monthly}, 101(8):715--724, 1994.

\bibitem{thimbleby1994displaying}
Harold~W. Thimbleby, Stuart Inglis, and Ian~H. Witten.
\newblock Displaying 3d images: Algorithms for single-image random-dot
  stereograms.
\newblock {\em Computer}, 27(10):38--48, 1994.

\bibitem{hearn2013}
{Tristan Hearn}.
\newblock magiceye-solver, 2013.
\newblock [Online; accessed 12-30-2020].

\bibitem{tyler1990autostereogram}
Christopher~W Tyler and Maureen~B Clarke.
\newblock Autostereogram.
\newblock In {\em Stereoscopic displays and applications}, volume 1256, pages
  182--197. International Society for Optics and Photonics, 1990.

\bibitem{ulyanov2016instance}
Dmitry Ulyanov, Andrea Vedaldi, and Victor Lempitsky.
\newblock Instance normalization: The missing ingredient for fast stylization.
\newblock {\em arXiv preprint arXiv:1607.08022}, 2016.

\bibitem{wang2004image}
Zhou Wang, Alan~C Bovik, Hamid~R Sheikh, and Eero~P Simoncelli.
\newblock Image quality assessment: from error visibility to structural
  similarity.
\newblock {\em IEEE transactions on image processing}, 13(4):600--612, 2004.

\bibitem{wikiautostereogram}
{Wikipedia contributors}.
\newblock Autostereogram, 2020.
\newblock [Online; accessed 12-30-2020].

\end{thebibliography}
}

\clearpage

\end{document}